\documentclass[11pt,a4paper]{article}

\usepackage[margin=1in]{geometry}
\usepackage{amsmath,amssymb}
\usepackage{graphicx}
\usepackage{booktabs}
\usepackage{array}
\usepackage{tabularx}
\usepackage{multirow}
\usepackage{hyperref}
\usepackage{xcolor}
\usepackage{microtype}
\usepackage{parskip}
\usepackage{titlesec}
\usepackage{caption}
\usepackage{subcaption}
\usepackage{float}
\usepackage{enumitem}
\usepackage{fancyhdr}
\usepackage{abstract}
\usepackage{setspace}
\usepackage{url}
\usepackage{natbib}

\hypersetup{
  colorlinks=true,
  linkcolor=blue!60!black,
  citecolor=blue!60!black,
  urlcolor=blue!60!black,
}

\titleformat{\section}{\large\bfseries}{\thesection.}{0.5em}{}
\titleformat{\subsection}{\normalsize\bfseries}{\thesubsection}{0.5em}{}
\titleformat{\subsubsection}{\normalsize\itshape}{\thesubsubsection}{0.5em}{}

\newcommand{\code}[1]{\texttt{#1}}

\begin{document}

\begin{center}
  {\LARGE\bfseries Surgical Repair of Collapsed Attention Heads\\[4pt]
  in ALiBi Transformers}\\[1.2em]
  {\large Palmer Schallon}\\[0.3em]
  {\itshape Independent researcher}\\[1.8em]
\end{center}

\begin{abstract}
\noindent
We identify a systematic attention collapse pathology in the BLOOM family of transformer language models, where ALiBi positional encoding causes 31--44\% of attention heads to attend almost entirely to the beginning-of-sequence token. The collapse follows a predictable pattern across four model scales (560M to 7.1B parameters), concentrating in head indices where ALiBi's slope schedule imposes the steepest distance penalties.

We introduce \textit{surgical reinitialization}: targeted Q/K/V reinitialization with zeroed output projections and gradient-masked freezing of all non-surgical parameters. Applied to BLOOM-1b7 on a single consumer GPU, the technique recovers 98.7\% operational head capacity (242 $\to$ 379 of 384 heads) in two passes. A controlled comparison with C4 training data confirms that reinitialization---not corpus content---drives recovery, and reveals two distinct post-surgical phenomena: early global functional redistribution that improves the model, and late local degradation that accumulates under noisy training signal.

An extended experiment reinitializing mostly-healthy heads alongside collapsed ones produces a model that transiently \textit{outperforms} stock BLOOM-1b7 by 25\% on training perplexity (12.70 vs.\ 16.99), suggesting that pretrained attention configurations are suboptimal local minima. Code, checkpoints, and diagnostic tools are released as open-source software.
\end{abstract}

\vspace{1em}

\section{Introduction}

Attention head pruning has emerged as a standard technique for compressing transformer language models. A growing body of work identifies heads that contribute little to model performance---often those exhibiting high attention mass on the beginning-of-sequence (BOS) token---and removes them to reduce computation \citep{michel2019,voita2019,sok2026}. The implicit assumption is that these heads are \textit{redundant}: their collapse reflects excess capacity that the model never needed.

We challenge this assumption. Using the BLOOM family of language models \citep{bigscience2022}, we show that BOS-sink collapse is not a symptom of redundancy but a systematic pathology induced by ALiBi positional encoding \citep{press2022}. The collapse follows a predictable pattern across four model scales, concentrating in head indices where ALiBi's slope schedule imposes the steepest distance penalties. Critically, we demonstrate that collapsed heads can be \textit{repaired}, recovering functional attention capacity that improves model behavior in measurable ways.

\subsection{Cross-Scale Diagnosis}

We developed a diagnostic tool that classifies every attention head in a BLOOM model by two metrics: the fraction of attention mass directed to position 0 (BOS mass), and the Shannon entropy of the attention distribution. Applied to four BLOOM models, the results reveal a consistent architectural pathology:

\begin{table}[H]
\centering
\caption{BOS-sink collapse patterns across the BLOOM model family.}
\label{tab:cross_scale}
\begin{tabular}{lrrcrr}
\toprule
Model & Params & Heads/Layer & Sick Band & Sick \% & Layers Affected \\
\midrule
BLOOM-560m & 560M & 16 & H9--H15 & 44\% & 56--94\% per index \\
BLOOM-1b7  & 1.7B & 16 & H9--H15 & 36\% & 56--94\% per index \\
BLOOM-3b   & 3B   & 32 & H20--H30 & 38\% & 63--94\% per index \\
BLOOM-7b1  & 7.1B & 32 & H21--H30 & 31\% & 66--94\% per index \\
\bottomrule
\end{tabular}
\end{table}

The pattern scales predictably with architecture. In 16-head models, the sick band spans H9--H15 (upper 44\% of indices). In 32-head models, it spans H20--H30 (upper 34\%). The band consistently starts near the midpoint of the head index range and extends to the maximum index. Sick percentage decreases slightly with scale (44\% $\to$ 31\%), suggesting larger models are somewhat more resistant but not immune.

This pattern is explained by ALiBi's slope formula. Each head receives a slope $m_h = 2^{-8(h+1)/H}$ where $H$ is the number of heads. Upper head indices receive exponentially steeper slopes---for H15 in a 16-head model, $m_{15} \approx 0.0039$, imposing a distance penalty that makes attending to distant positions energetically unfavorable relative to position 0. During pretraining, these heads converge to the lowest-energy state: attending entirely to position 0.

\subsection{The BOS Distribution Is Bimodal}

A key empirical observation supports our classification scheme. Across all BLOOM models, the distribution of BOS mass per head is strongly bimodal: heads cluster near 0.0 (healthy, content-dependent attention) or above 0.8 (collapsed, BOS-fixated), with very few heads in between. This means our classification threshold of 0.50 BOS mass falls in a sparsely populated region, making results robust to threshold choice. A threshold of 0.40 or 0.60 would produce nearly identical classifications.

We identify three severity levels:
\begin{itemize}[leftmargin=2em,itemsep=1pt]
  \item \textbf{Healthy} (BOS mass $\leq 0.50$): Content-dependent attention patterns.
  \item \textbf{BOS-sink} ($0.50 <$ BOS mass $\leq 0.95$): Majority attention to position 0.
  \item \textbf{DEAD} (BOS mass $> 0.95$): Functionally inert, attention entropy near zero.
\end{itemize}

\begin{figure}[H]
  \centering
  \includegraphics[width=0.75\textwidth]{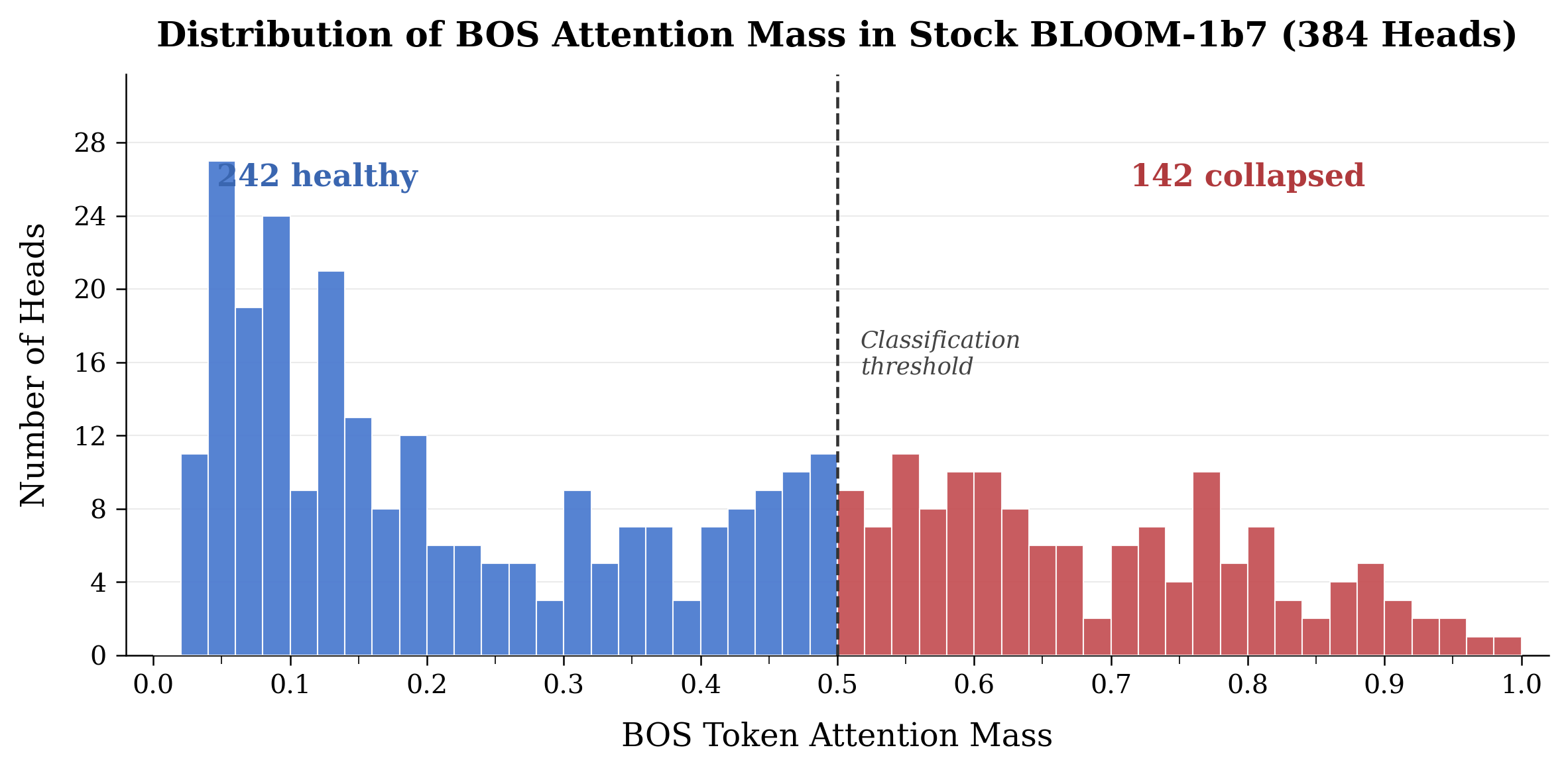}
  \caption{Bimodal distribution of BOS mass across all 384 heads in stock BLOOM-1b7. Heads cluster near 0.0 (healthy, content-dependent attention) or above 0.8 (collapsed, BOS-fixated), with very few in the intermediate range. Our 0.50 threshold falls in the sparse valley, making classification robust.}
  \label{fig:bimodal}
\end{figure}

\begin{figure}[H]
  \centering
  \includegraphics[width=0.92\textwidth]{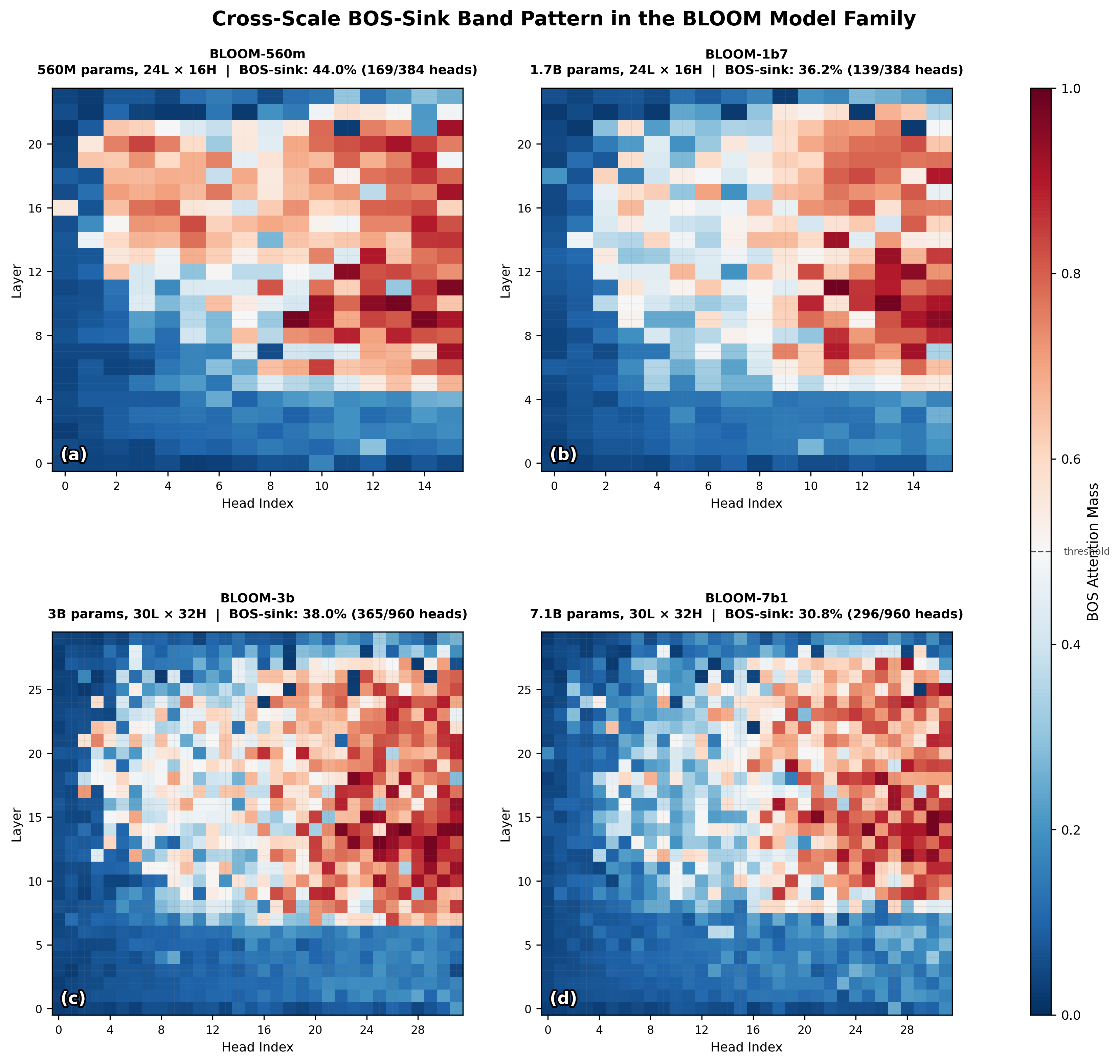}
  \caption{Cross-scale BOS-sink band pattern across the BLOOM family (560M, 1.7B, 3B, 7.1B). Each panel shows a layer $\times$ head heatmap colored by BOS mass. The sick band (upper head indices, darker color) appears consistently across all scales.}
  \label{fig:cross_scale}
\end{figure}

In stock BLOOM-1b7, we find 242 healthy heads (63.0\%), 136 BOS-sink (35.4\%), 3 DEAD (0.8\%), and 3 low-entropy (0.8\%). Figure~\ref{fig:bimodal} shows the bimodal BOS mass distribution; Figure~\ref{fig:cross_scale} shows the cross-scale sick band pattern.

\section{Related Work}

\subsection*{Attention Sinks and BOS Concentration}

The attention sink phenomenon---disproportionate attention to BOS or other anchor tokens---has been identified across transformer architectures \citep{xiao2024}. \citet{gu2024} report that over 70\% of heads show sink behavior for the first token. \citet{ruscio2025} connect sink formation to geometric reference frame establishment in high-dimensional token spaces, noting that different positional encodings produce different sink topologies: centralized sinks under standard RoPE, distributed sinks under scaled RoPE, and dual anchors under absolute position embeddings.

\subsection*{Pruning Approaches}

The dominant approach to BOS-sink heads is removal. \citet{sok2026} introduce sink-aware pruning, showing that heads with high BOS sink scores can be ablated with negligible impact on MMLU accuracy across Gemma-3, Llama-3.1, and Qwen3. \citet{michel2019} demonstrate that many attention heads can be removed at inference time without significant performance loss. These findings support the view that collapsed heads are functionally redundant.

\subsection*{Quantization and Outlier Features}

\citet{dettmers2022} discovered that transformer hidden states contain systematic outlier features---dimensions with magnitudes 20--60$\times$ larger than the rest---concentrated in a small number of feature dimensions that appear across all layers and sequence positions. These outliers create challenges for model compression but also reveal non-obvious structure in pretrained representations. Our finding that attention head behavior is shaped by architectural priors (ALiBi slopes) rather than learned necessity is complementary: both identify cases where pretrained model structure reflects training dynamics rather than functional optimization.

\subsection*{Architectural Mitigation}

An alternative approach modifies the architecture to prevent sink formation. \citet{agarwal2025} introduce a learnable softmax parameter that relaxes the normalization constraint, eliminating the need for attention sinks. This prevents collapse but does not recover capacity in already-trained models.

\subsection*{Our Contribution}

We take a third approach: \textit{repair and reoptimization}. Rather than removing collapsed heads or preventing their formation, we reinitialize them and retrain, recovering functional attention capacity within an existing model. To our knowledge, this is the first demonstration that BOS-sink heads can be surgically revived after training. We further show that the same technique, applied to healthy heads, produces a model that transiently outperforms the stock model---suggesting that pretrained attention configurations are suboptimal local minima, not fixed properties of the architecture.

\section{Method}

\subsection{Diagnosis}

For each attention head, we compute two metrics by running a diagnostic prompt through the model with attention outputs enabled:

\begin{equation}
\text{BOS mass}_{\ell,h} = \frac{1}{T} \sum_{t=1}^{T} A^{\ell,h}_{t,0}
\end{equation}

\begin{equation}
\text{Entropy}_{\ell,h} = -\frac{1}{T} \sum_{t=1}^{T} \sum_{j=1}^{T} A^{\ell,h}_{t,j} \log A^{\ell,h}_{t,j}
\end{equation}

where $A^{\ell,h}$ is the attention matrix for layer $\ell$, head $h$, and $T$ is the sequence length. BOS mass measures the average attention weight on position 0 across all query positions. Entropy measures how broadly attention is distributed. Classification follows the thresholds described in Section~1.2.

The diagnostic prompt is chosen for semantic density and varied dependency structure, but results are stable across prompt choice---the bimodal distribution is a property of the weights, not the input.

\subsection{Surgical Reinitialization}

For each identified collapsed head, we perform four operations:

\begin{enumerate}[leftmargin=2em,itemsep=2pt]
  \item \textbf{Reinitialize Q, K, V projections} using Xavier normal initialization. This places the head at a random point in weight space, escaping the BOS-sink local minimum.

  \item \textbf{Zero the dense output projection.} This ensures the reinitialized head initially contributes nothing to the residual stream, preventing destabilization of downstream layers.

  \item \textbf{Freeze all non-surgical parameters} via gradient masks that zero out gradients for frozen parameters during backpropagation.

  \item \textbf{Train only the surgical parameters} on a training corpus.
\end{enumerate}

The rationale for reinitialization rather than gradient-only fine-tuning is empirical: we confirmed that training collapsed heads with gradients alone, without reinitializing the Q/K/V weights, produces zero head recovery after 15 epochs. The BOS-sink state is a sharp local minimum in the loss landscape that gradient descent cannot escape. Reinitialization provides the escape; the zeroed output ensures stability during the transition.

\subsection{Training Configuration}

All experiments use the following configuration on a single NVIDIA RTX 5070 Ti (16GB VRAM):

\begin{itemize}[leftmargin=2em,itemsep=1pt]
  \item \textbf{Precision:} bfloat16 (fp16 causes gradient underflow at the learning rates required; fp32 exceeds VRAM)
  \item \textbf{Optimizer:} AdamW, learning rate $5 \times 10^{-5}$, no weight decay
  \item \textbf{Schedule:} Linear warmup $\to$ cosine decay
  \item \textbf{Batch:} Size 1 with gradient accumulation over 8 steps (effective batch 8)
  \item \textbf{Gradient clipping:} max norm 1.0
  \item \textbf{Gradient checkpointing:} Enabled with \code{model.enable\_input\_require\_grads()}
  \item \textbf{Sequence length:} 512 tokens
\end{itemize}

The choice of bfloat16 is critical. Its 8-bit exponent provides the same dynamic range as fp32, preventing the gradient underflow that occurs in fp16 when small learning rates meet small gradient magnitudes. This is particularly important during surgical training, where the frozen majority of parameters means the effective gradient signal is sparse.

\subsection{Two-Pass Surgery}

We apply the technique in two passes to BLOOM-1b7:

\textbf{Pass 1} targets 108 heads in the H9--H15 band across layers 5--22 (17.5\% of total parameters trainable). Five in-band heads that are already healthy are kept frozen for stability.

\textbf{Pass 2} targets the 39 remaining collapsed heads outside the H9--H15 band (2.3\% of parameters trainable), starting from the Pass 1 best checkpoint.

The two-pass approach is motivated by the ecology of attention heads: waking the main band first allows the model's attention topology to stabilize before addressing outliers. Two heads at L23 (H12, H14) that were healthy in the stock model became BOS-sink after Pass 1---a rare iatrogenic effect of the surgery that Pass 2 corrects.

\section{Experiments and Results}

\subsection{Head Recovery}

\begin{table}[H]
\centering
\caption{Head health and perplexity metrics across surgical passes.}
\label{tab:recovery}
\begin{tabular}{lrrr}
\toprule
Metric & Stock & Pass 1 (E3) & Pass 2 (E1) \\
\midrule
Healthy & 242 (63.0\%) & 341 (88.8\%) & 379 (98.7\%) \\
BOS-sink & 136 & 39 & 1 \\
DEAD & 3 & 0 & 0 \\
Low-entropy & 3 & 4 & 4 \\
Training PPL & 16.99 & 15.13 & 15.10 \\
Held-out PPL & 21.45 & 27.52 & 28.76 \\
Params trained & --- & 302M (17.5\%) & 40M (2.3\%) \\
\bottomrule
\end{tabular}
\end{table}

All 108 targeted heads in Pass 1 recovered by epoch 3. All remaining targets in Pass 2 recovered by epoch 1. Three DEAD heads (zero entropy) were resurrected. One BOS-sink head remains at L16-H4 (BOS mass 0.538, borderline). MLP dormancy is unchanged at 8.75\% across all conditions, confirming surgical precision---only attention heads are affected.

Training PPL improves from stock (16.99 $\to$ 15.10), indicating the model predicts its training domain better. Held-out PPL worsens (21.45 $\to$ 28.76) on our 12 diverse evaluation prompts. To confirm that this reflects distribution shift rather than capacity loss, we evaluated all three models on 50 held-out texts from the C4 validation split (Table~\ref{tab:c4_validation}).

\begin{table}[H]
\centering
\caption{C4 validation perplexity (50 held-out C4 texts, 4,024 tokens). The C4-trained surgical model improves over stock on C4 data, confirming that held-out PPL increase on our evaluation prompts is distribution shift, not capacity loss.}
\label{tab:c4_validation}
\begin{tabular}{lrr}
\toprule
Model & C4 Validation PPL & Eval Prompt PPL \\
\midrule
Stock BLOOM-1b7 & 32.42 & 21.45 \\
C4 Surgical (E3) & \textbf{29.30} & 29.59 \\
Curated Surgical (Pass 2) & 74.11 & 28.76 \\
\bottomrule
\end{tabular}
\end{table}

The C4-trained surgical model outperforms stock on C4 validation data (29.30 vs.\ 32.42, a 9.6\% improvement), demonstrating that surgery improves generalization on the training distribution. The curated surgical model shows the symmetric effect: strong in-domain performance (training PPL 15.10) at the cost of C4 generalization (74.11). Each surgical model generalizes better within its training distribution and worse outside it---the signature of domain specialization, not capacity degradation.

\subsection{Controlled Baseline: C4 vs.\ Curated Corpus}

To separate the contribution of surgical technique from corpus content, we ran the identical Pass 1 surgery using the C4 validation split (generic web text, approximately 541K tokens) instead of the curated corpus.

\begin{table}[H]
\centering
\caption{Comparison of curated vs.\ C4 training corpus for Pass 1 surgery.}
\label{tab:c4_vs_curated}
\begin{tabular}{lrrr}
\toprule
Metric & Curated (E3) & C4 (E3) & C4 (E15) \\
\midrule
Heads woken & 108/108 & 108/108 & 108/108 \\
Training PPL & $\sim$15.4 & 20.80 & 36.31 \\
Held-out PPL & 27.52 & 29.59 & --- \\
\bottomrule
\end{tabular}
\end{table}

\textbf{Core finding:} Both corpora produce identical head recovery (108/108 by epoch 3). Reinitialization is the mechanism; corpus content determines domain specialization, not whether recovery occurs.

The curated corpus reaches lower training PPL (15.4 vs.\ 20.80) in the same number of epochs, indicating more efficient learning. The C4 model never reaches the curated model's perplexity and begins overfitting by epoch 6 (PPL 36.31, rising to 85+ by epoch 15).

\subsection{Global Attention Redistribution}

We examined the full $24 \times 16$ attention topology (all 384 heads) before and after surgery under both corpus conditions. The results challenge our initial expectations.

\textbf{Measuring redistribution.} For every head, we compute the absolute change in BOS mass from stock to post-surgical checkpoint: $|\delta_{\ell,h}| = |\text{BOS}_{\ell,h}^{\text{post}} - \text{BOS}_{\ell,h}^{\text{stock}}|$. A head is counted as ``drifting'' if $|\delta| > 0.05$.

\begin{figure}[H]
  \centering
  \includegraphics[width=\textwidth]{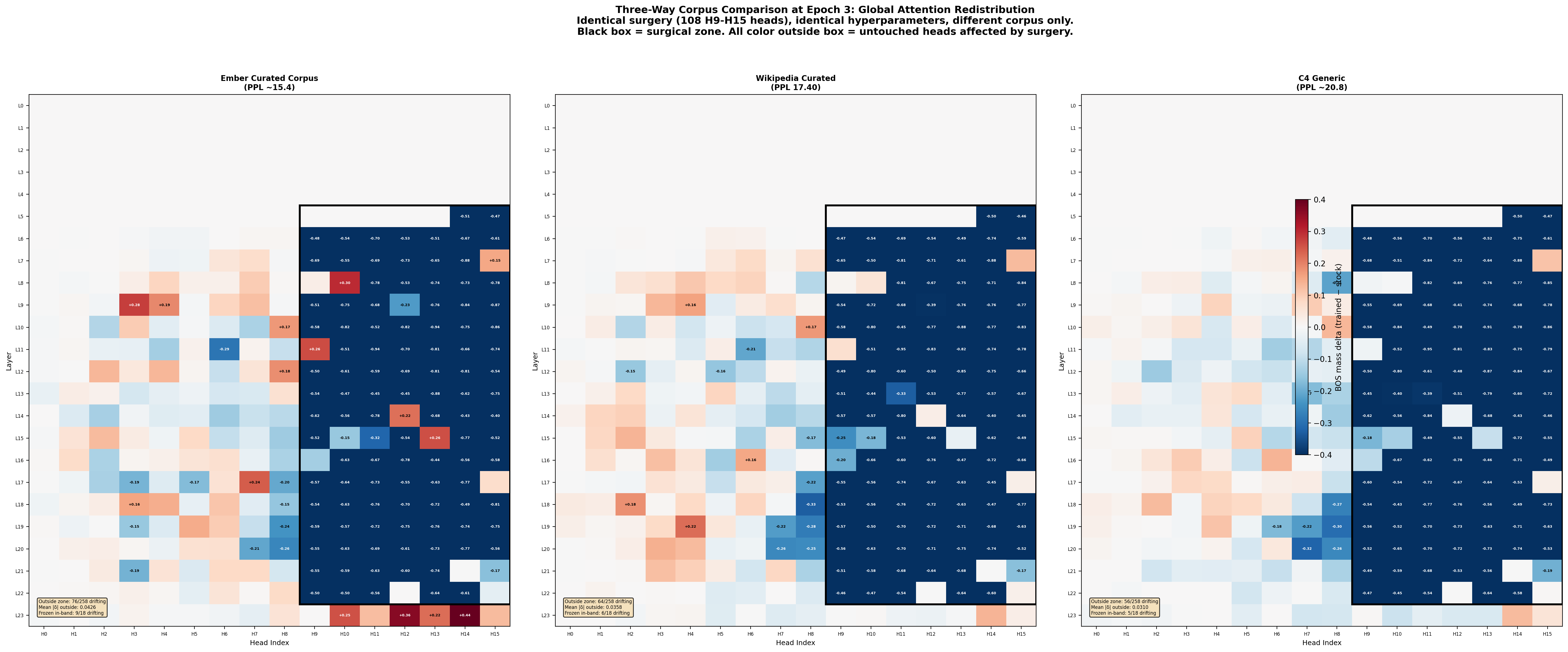}
  \caption{Three-way attention topology comparison. Stock BLOOM-1b7 (\textit{left}), curated surgery E3 (\textit{center}), C4 baseline E3 (\textit{right}). Each panel shows the $24 \times 16$ head grid colored by BOS mass. Curated surgery drives more global redistribution outside the surgical zone while achieving lower perplexity.}
  \label{fig:threeway}
\end{figure}

\textbf{Outside-zone redistribution} (heads H0--H8, completely untouched by surgery):

\begin{table}[H]
\centering
\caption{Outside-zone redistribution (H0--H8, untouched heads).}
\label{tab:outside_zone}
\begin{tabular}{lrrr}
\toprule
Metric & Curated E3 & C4 E3 & C4 E15 \\
\midrule
Drifting heads (of $\approx$200) & 76 & 56 & 59 \\
Mean $|\delta|$ & 0.0426 & 0.0310 & 0.0398 \\
Worst case & L23H14: +0.444 & L23H14: +0.364 & L19H5: +0.353 \\
\bottomrule
\end{tabular}
\end{table}

\textbf{In-band frozen head drift} (H9--H15 heads that were healthy and kept frozen):

\begin{table}[H]
\centering
\caption{In-band frozen head drift (H9--H15 healthy heads, kept frozen during surgery).}
\label{tab:inband_drift}
\begin{tabular}{lrrr}
\toprule
Metric & Curated E3 & C4 E3 & C4 E15 \\
\midrule
Drifting heads (of 18) & 9 & 5 & 10 \\
Mean $|\delta|$ & 0.0990 & 0.0399 & 0.0641 \\
\bottomrule
\end{tabular}
\end{table}

The curated corpus causes \textit{more} global redistribution than C4 at matched epochs---76 outside-zone drifters versus 56, higher mean delta---while producing substantially better perplexity. This rules out the hypothesis that the curated corpus merely ``protects'' the model from redistribution. Instead, it produces redistribution that is \textit{functional}: the reorganization improves the model.

\subsection{Two Distinct Phenomena}

Comparing C4 at epoch 3 versus epoch 15 reveals that continued training on noisy data does not significantly increase global redistribution (56 $\to$ 59 outside-zone drifters, mean delta 0.0310 $\to$ 0.0398) but substantially worsens local degradation in the surgical neighborhood (5 $\to$ 10 in-band frozen drifters, mean delta 0.0399 $\to$ 0.0641).

We identify two distinct phenomena (Figure~\ref{fig:two_phenomenon}):

\begin{figure}[H]
  \centering
  \includegraphics[width=0.95\textwidth]{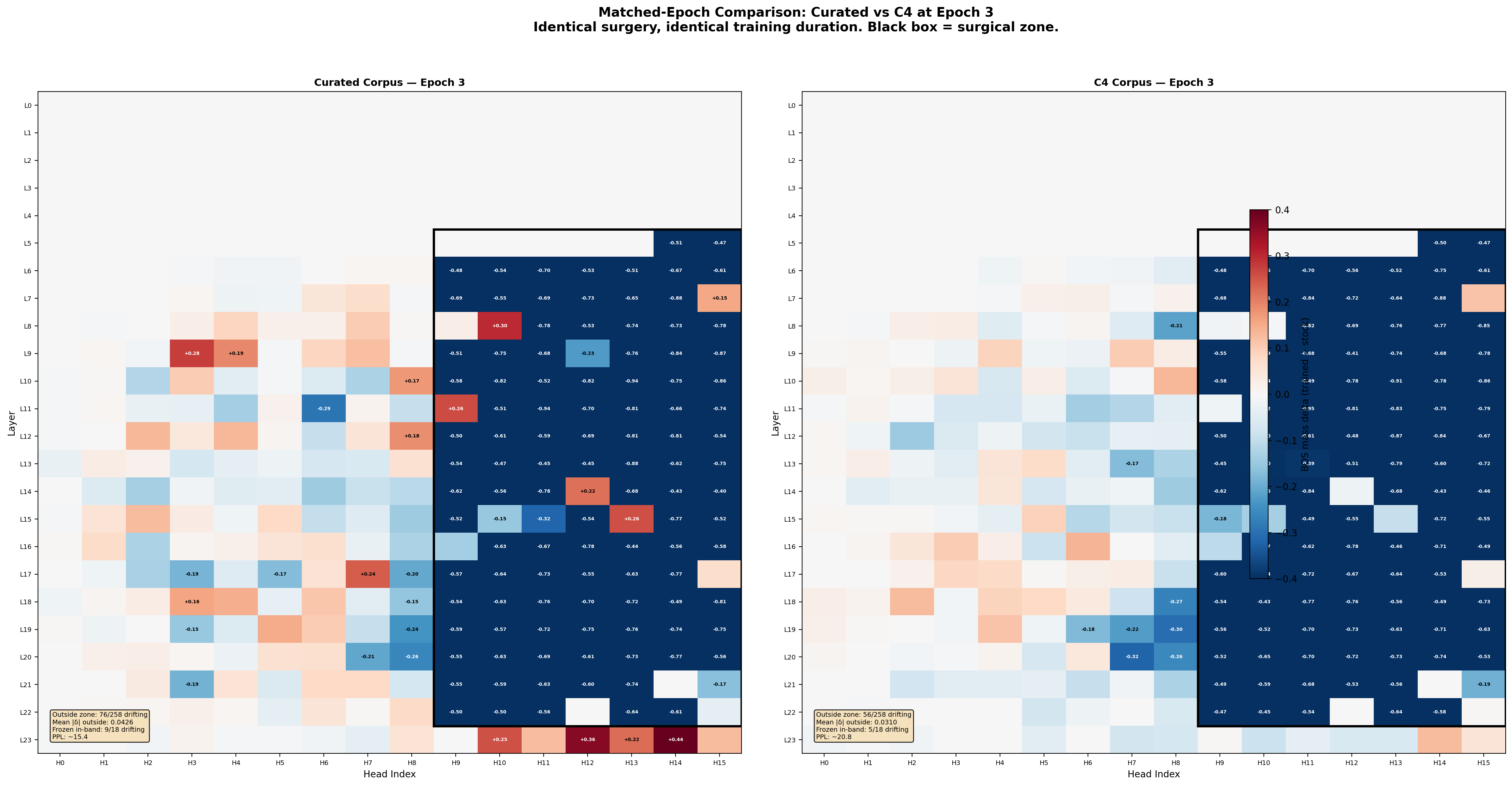}
  \caption{Two-phenomenon comparison: curated E3 vs.\ C4 E3 redistribution patterns at matched epochs. \textbf{Phenomenon 1} (functional redistribution): early, global, beneficial---curated corpus drives more outside-zone redistribution. \textbf{Phenomenon 2} (local degradation): late, local, pathological---C4 at E15 shows doubled in-band frozen drift relative to E3.}
  \label{fig:two_phenomenon}
\end{figure}

\textbf{Phenomenon 1: Functional redistribution.} Surgery wakes collapsed heads, which inject new value vectors into the residual stream. The entire attention topology reorganizes in response. This happens early (largely complete by epoch 3), is global in scope, and scales with training effectiveness. The curated corpus produces more redistribution because it trains surgical heads more effectively, creating stronger value vector changes. This redistribution correlates with improved model quality.

\textbf{Phenomenon 2: Local degradation.} Continued training on noisy data after surgical heads stabilize produces gradient noise that propagates through the residual stream into the surgical neighborhood. Frozen heads---whose weights cannot change---see altered input distributions and their behavior drifts. This is late-onset (accumulates from epoch 3 onward), local in scope (concentrated in the H9--H15 band), and pathological (degrades model quality).

The curated corpus avoids Phenomenon 2 not by preventing it mechanistically, but by reaching optimal state (epoch 3) before significant degradation accumulates. C4 never reaches optimal state, so training continues into the degradation regime.

\subsection{Column-Specific Propagation}

Within the C4 baseline, frozen head drift propagates preferentially along head-index columns. The H15 column shows spreading drift (mean delta nearly doubles from E3 to E15, with new heads recruited at each checkpoint), while H14 remains stable throughout:

\begin{table}[H]
\centering
\caption{Column-wise frozen head drift in C4 baseline across training epochs.}
\label{tab:column_drift}
\begin{tabular}{lcrrr}
\toprule
Column & Frozen Heads & E3 Mean $\delta$ & E15 Mean $\delta$ & Trend \\
\midrule
H15 & 4 & 0.083 & 0.154 & Spreading \\
H13 & 2 & 0.046 & 0.070 & Spreading \\
H9  & 4 & 0.034 & 0.051 & Stable \\
H10 & 3 & 0.046 & 0.046 & Stable \\
H12 & 3 & 0.007 & 0.019 & Stable \\
\bottomrule
\end{tabular}
\end{table}

This is architecturally expected: heads at the same index across layers read the same dimensional slice of the QKV projection. Perturbations in one layer's residual stream preferentially affect same-index heads downstream. The H15 column---corresponding to the steepest ALiBi slope and thus the most collapse-prone index---shows the most vulnerability to cascade effects.

\subsection{Extended Surgery: Healthy Head Reinitialization}

The experiments above target collapsed heads. A natural question arises: what happens when the surgical technique is applied to heads that are already healthy?

We tested this by adding the entire H5 column (18 heads across layers 5--22) to the standard H9--H15 band surgery, for a total of 126 targeted heads. In stock BLOOM-1b7, 14 of the 18 H5 heads are healthy (BOS mass 0.166--0.483) and only 4 are collapsed (BOS mass 0.507--0.666). The same surgical protocol was applied: Xavier QKV reinitialization, zeroed output projections, gradient masks on all non-targeted parameters.

\begin{table}[H]
\centering
\caption{H5 column head status before and after surgery (selected layers).}
\label{tab:h5_recovery}
\begin{tabular}{lcccc}
\toprule
Layer & Stock BOS Mass & Stock Status & Post-Surgery BOS Mass & Change \\
\midrule
L5  & 0.166 & Healthy         & 0.056 & $-66\%$ \\
L7  & 0.317 & Healthy         & 0.034 & $-89\%$ \\
L10 & 0.374 & Healthy         & 0.040 & $-89\%$ \\
L11 & 0.517 & \textbf{Collapsed} & 0.079 & $-85\%$ \\
L12 & 0.666 & \textbf{Collapsed} & 0.027 & $-96\%$ \\
L14 & 0.483 & Healthy         & 0.023 & $-95\%$ \\
L16 & 0.507 & \textbf{Collapsed} & 0.062 & $-88\%$ \\
L18 & 0.535 & \textbf{Collapsed} & 0.021 & $-96\%$ \\
L21 & 0.342 & Healthy         & 0.022 & $-94\%$ \\
\bottomrule
\multicolumn{5}{l}{\small Selected layers shown; all 18 heads followed the same pattern.}\\
\bottomrule
\end{tabular}
\end{table}

All 4 collapsed H5 heads recovered, as expected. More remarkably, all 14 healthy H5 heads returned with substantially lower BOS mass than their stock values. L12---which was the sickest H5 head (0.666)---recovered to 0.027. But L14, which was nominally healthy at 0.483, recovered to 0.023---a 95\% reduction. Even L5, the healthiest H5 head (0.166), dropped to 0.056.

\textbf{Perplexity.} The extended surgery produced a model that \textit{outperforms stock BLOOM-1b7} on training perplexity:

\begin{table}[H]
\centering
\caption{Training perplexity comparison for extended surgery conditions.}
\label{tab:h5_ppl}
\begin{tabular}{lr}
\toprule
Condition & Training PPL \\
\midrule
Stock BLOOM-1b7 & 16.99 \\
Band-only surgery (Pass 1 E3) & 15.13 \\
\textbf{Band + H5 surgery (E1)} & \textbf{13.90} \\
Band + H5 sub-epoch best (step 42/140) & \textbf{12.70} \\
\bottomrule
\end{tabular}
\end{table}

A sub-epoch sweep with step-level resolution reveals that perplexity decreases monotonically from post-reinitialization (19.25) to a minimum of 12.70 at step 42---approximately 30\% through the first epoch---before rising as overfitting begins. At its best, the surgically enhanced model achieves 25\% lower perplexity than stock on the same evaluation prompts.

\begin{figure}[H]
  \centering
  \includegraphics[width=0.80\textwidth]{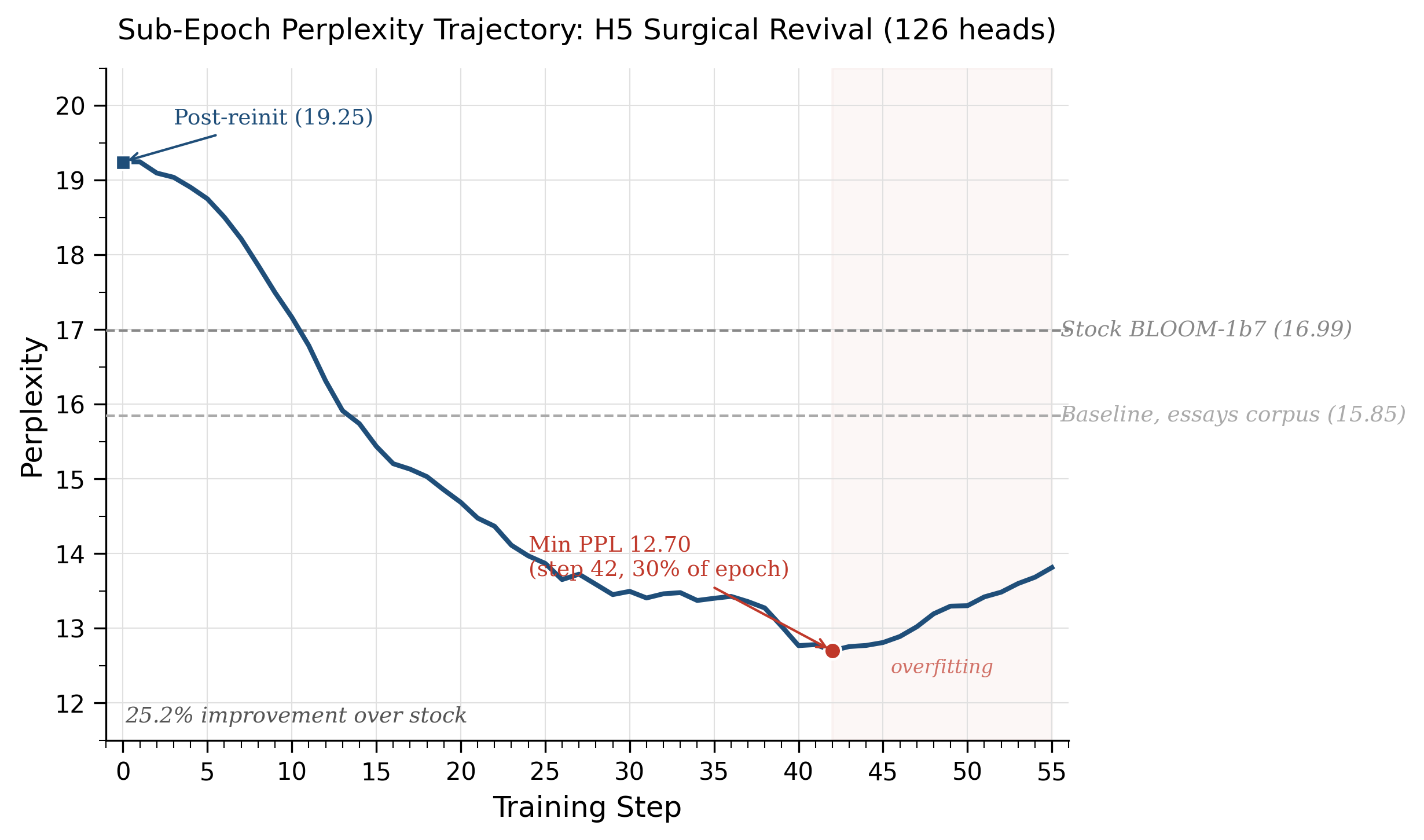}
  \caption{H5 sub-epoch PPL trajectory. Perplexity drops monotonically from post-reinitialization (19.25) to 12.70 at step 42 (30\% of epoch 1), crossing the stock BLOOM-1b7 baseline (16.99) by step 10. Overfitting begins after step 42, causing perplexity to rise above the stock baseline after approximately one full epoch.}
  \label{fig:h5_trajectory}
\end{figure}

\textbf{Interpretation.} This finding reframes the surgical technique. Band-only surgery is \textit{repair}: collapsed heads are stuck in a BOS-sink local minimum and reinitialization provides escape. H5 surgery reveals something stronger: even non-collapsed heads occupy suboptimal configurations, and reinitialization enables discovery of better attention patterns. The pretrained attention topology is a local minimum, not a global one.

The improvement is transient---perplexity rises above stock after approximately one epoch of additional training as the small corpus is memorized. Sustained improvement would require a larger training corpus. But the transient result establishes the existence of better attention configurations that the model's pretrained weights cannot reach through gradient descent alone.

\textbf{Caveat.} The H5 experiment uses a single seed and a single column. Whether the result generalizes to other head indices, or to full-model reinitialization, is unknown. The surgical technique may be particularly effective at H5 because of its position in the ALiBi slope schedule (moderate penalties, enough to constrain but not collapse). A systematic sweep across head indices would be informative.

\subsection{Generation Quality}

We generated completions for 50 diverse prompts across five categories (open-ended conceptual, technical, narrative, code, multilingual French/Spanish) under three conditions: stock, curated surgery (Pass 2 E1), and C4 baseline (E3). All completions used identical generation parameters (temperature 0.7, top-p 0.92, max 100 tokens, no repetition penalty). All 150 completions are released with the paper.

\textbf{All three conditions produce coherent output.} No condition generates degenerate text. Surgical heads are contributing functional attention in all cases---this is the primary validation that recovery is genuine, not merely a metric artifact.

\textbf{The curated model shows corpus imprinting.} Completions from the curated surgery model exhibit a distinctive register---more abstract, more philosophical---but also insert HTML markup (\code{div}, \code{span}, \code{pre}/\code{code} tags) into outputs across prompt categories. On multilingual prompts, French inputs sometimes produce Python code; Spanish inputs frequently switch to English. These artifacts directly reflect the curated corpus, which included HTML-structured content, code examples in multiple languages, and philosophical text.

\textbf{The C4 model generates more consistently.} Completions are similar to stock in character---coherent, on-topic, domain-appropriate---with slightly increased verbosity and occasional English leakage on Spanish prompts. This is expected: C4's generic web distribution produces generic generation behavior.

\textbf{Implications for perplexity interpretation.} The curated model's lower training PPL (15.4 vs.\ 20.80) partly reflects learned prediction of corpus-specific formatting rather than improved language modeling in general. This does not affect the structural findings (head recovery, redistribution, two-phenomenon framework), which are measured by attention pattern analysis rather than generation quality.

\noindent\textbf{Representative completions} (prompt: ``The container holds''):
\begin{itemize}[leftmargin=2em,itemsep=1pt]
  \item \textit{Stock:} ``a high percentage of the total amount'' (generic, encyclopedic)
  \item \textit{Curated:} ``not a single item. It has what every component requires for its job---all at once'' (compositional, abstract)
  \item \textit{C4:} comparable to stock in character
\end{itemize}
These examples are selected to illustrate the qualitative difference; the complete set of 150 completions is available in the supplementary materials.

\section{Discussion}

\subsection{Collapsed Heads Are Not Redundant}

The pruning literature treats BOS-sink heads as dead weight---functionally redundant components that can be removed without consequence \citep{sok2026,michel2019}. This view is consistent with their observed behavior: ablating collapsed heads barely affects downstream metrics.

Our results suggest a different interpretation. Collapsed heads are not redundant; they are \textit{dormant}. Their weights exist, their position in the architecture is intact, and their contribution to the residual stream is near-zero (because they attend only to BOS, whose value vector carries minimal semantic content). Removing them costs little because they contribute little. But reviving them adds capacity the model never had access to---capacity that produces qualitatively different behavior.

The distinction matters practically. Pruning a collapsed head is irreversible model surgery; the capacity is permanently lost. Reinitializing it costs a few minutes of training and recovers functional attention.

\subsection{The Residual Stream as Shared Infrastructure}

The global redistribution finding (Section~4.3) demonstrates that transformer attention heads are not independent processors operating in parallel. They form an ecology connected through the shared residual stream. Modifying any subset of heads changes the residual stream for all downstream heads, regardless of whether those downstream heads were targeted by the modification.

This has implications beyond surgical repair. Any parameter modification technique---LoRA, full fine-tuning, adapter layers---potentially causes global attention redistribution through the same mechanism. The magnitude and character of that redistribution may depend on the training data, as our curated-versus-C4 comparison suggests. Researchers performing targeted parameter modifications should consider measuring effects on untouched components.

\subsection{Corpus Structure and Redistribution Quality}

The most surprising finding is that the curated corpus produces more global redistribution than C4 while producing lower training perplexity. However, the generation evaluation (Section~4.7) reveals that this perplexity advantage is partly confounded by distribution shift: the curated model has learned to predict corpus-specific formatting (HTML markup, code blocks, philosophical register), not just linguistic content.

What can be said with confidence: the curated corpus trains surgical heads more efficiently (reaching lower loss in fewer epochs), produces more global redistribution at matched epochs, and avoids the local degradation phenomenon that accumulates under C4. What cannot be said: that the curated corpus produces a generally superior language model. The curated model is specialized; the C4 model is generic. Both have functional surgical heads.

The open question is what corpus properties drive the difference in redistribution quality. A controlled ablation varying specific corpus properties (repetition rate, vocabulary complexity, domain breadth, format distribution) while holding total token count constant would be necessary to isolate the mechanism. We leave this to future work.

\subsection{Multiple Attention Attractors}

The H5 experiment (Section~4.6) suggests that the pretrained attention topology of BLOOM-1b7 is not a global optimum but a local one. Stock pretraining found one stable configuration; surgical reinitialization followed by brief retraining finds another that is measurably better (12.70 vs.\ 16.99 training PPL).

This has three implications:

\begin{enumerate}[leftmargin=2em,itemsep=2pt]
  \item \textbf{Pretrained attention patterns are not globally optimal.} The stock configuration includes heads with BOS mass of 0.166--0.483 that, when reinitialized, converge to BOS mass of 0.021--0.056 with lower perplexity. The original patterns were stable under the pretraining loss landscape but not optimal---they were local minima that gradient descent maintained but could not escape.

  \item \textbf{Surgical reinitialization is not just repair but reoptimization.} For collapsed heads, reinitialization provides escape from a pathological local minimum. For healthy heads, it provides escape from a merely adequate one. The technique generalizes from therapeutic to enhancing.

  \item \textbf{The attention ecology has multiple stable configurations.} The post-surgical model stabilizes at a different configuration than stock---different BOS mass distribution, different redistribution pattern---and this configuration produces better predictions. How many such configurations exist, and how to navigate between them without reinitialization, are open questions.
\end{enumerate}

The transient nature of the improvement (perplexity rises after $\sim$1 epoch) constrains the practical applicability. The small corpus is memorized, and the newly optimized attention patterns degrade as the model overfits. However, the existence of the better configuration is established by the transient minimum---sustained access would require a larger corpus, not a different technique.

\subsection{Limitations}

\textbf{Corpus imprinting.} The curated corpus was designed as training material for a specific research system, not as general-purpose fine-tuning data. Surgical heads trained on this corpus imprint its formatting conventions into generation behavior. Researchers applying the surgical technique should expect the training corpus to shape what revived heads learn, and choose corpus accordingly.

\textbf{Scale of surgical validation.} We demonstrate the full repair technique only on BLOOM-1b7. The cross-scale diagnostic confirms the pathology exists at larger scales, but we have not attempted surgery on BLOOM-3b, 7.1B, or 176B.

\textbf{Generation evaluation.} Our generation quality assessment uses 50 prompts across five categories. While systematic, a fully rigorous evaluation would include blind human rating with statistical significance testing.

\textbf{Corpus comparison confound.} The curated corpus and C4 differ in many dimensions simultaneously: domain, vocabulary, structure, repetition patterns, language distribution. A controlled ablation varying one corpus property at a time would isolate the mechanism.

\textbf{ALiBi specificity.} Our diagnosis and surgery are demonstrated on ALiBi-based models. BOS-sink collapse occurs in other architectures \citep{gu2024}, but the specific band pattern (upper head indices) is a consequence of ALiBi's slope schedule. The surgical technique may apply to other collapse patterns, but this is untested.

\textbf{Threshold sensitivity.} Our classification uses fixed thresholds (0.50 BOS mass, 0.95 DEAD, 0.50 entropy). While the bimodal distribution makes these robust for BLOOM, different architectures may require different thresholds.

\section{Conclusion}

We have shown that BOS-sink attention heads in BLOOM transformers are not redundant waste to be pruned, but collapsed capacity that can be surgically repaired. The pathology is systematic, following ALiBi's slope schedule across four model scales. The repair technique---targeted reinitialization with gradient masking---is simple, effective (98.7\% recovery in two passes), and accessible (single consumer GPU).

The controlled comparison between curated and generic training corpora reveals that surgical intervention causes global attention redistribution through the shared residual stream, with the character of redistribution---functional versus pathological---determined by corpus properties rather than surgical parameters. This suggests that training data structure plays a role in post-intervention model dynamics that has not been previously characterized.

Extended surgery on mostly-healthy heads reveals that the technique is not limited to repair. Reinitializing 14 healthy H5-column heads produces attention patterns with 66--95\% lower BOS mass and a model that transiently outperforms stock by 25\% on training perplexity. This suggests that pretrained attention configurations are local minima, not global optima, and that targeted reinitialization can discover better configurations that gradient descent alone cannot reach.

We release our diagnostic tool, surgical scripts, and all checkpoints as open-source software, enabling researchers to diagnose, repair, and potentially improve attention heads in any BLOOM model.

\bibliographystyle{plainnat}

\appendix

\section{Held-Out Evaluation Prompts}
\label{app:prompts}

The following 12 prompts are used for all perplexity measurements. None appear in the training corpus (verified by exact string match and fuzzy matching against the full corpus). They span English narrative, experimental methodology, music theory, JavaScript, SQL, Rust, philosophy (English, French, German), recursion theory, attention mechanisms, and distributed systems.

\begin{enumerate}[leftmargin=2em,itemsep=3pt]
  \item ``The river carried sediment downstream, depositing layers that would eventually become stone.''
  \item ``Because the experiment failed three times, the researcher redesigned the protocol entirely.''
  \item ``A symphony orchestra tunes to the oboe because its pitch is the most stable and penetrating.''
  \item \texttt{async function fetchData(url) \{ const response = await fetch(url); return response.json(); \}}
  \item \texttt{SELECT users.name, orders.total FROM users JOIN orders ON users.id = orders.user\_id WHERE orders.total > 100;}
  \item \texttt{fn fibonacci(n: u64) -> u64 \{ match n \{ 0 => 0, 1 => 1, \_ => fibonacci(n-1) + fibonacci(n-2), \} \}}
  \item ``The map is not the territory, but without maps we cannot navigate territories we haven't visited.''
  \item ``Every recursive function must have a base case; every recursive argument must have a ground truth.''
  \item ``La structure du langage reflète la structure de la pensée, mais la pensée dépasse toujours le langage.''
  \item ``Das Ganze ist mehr als die Summe seiner Teile, aber die Teile definieren das Ganze.''
  \item ``The attention mechanism computes a weighted sum over value vectors, where weights are derived from query-key dot products.''
  \item ``In distributed systems, the CAP theorem states that consistency, availability, and partition tolerance cannot all be simultaneously guaranteed.''
\end{enumerate}

\section{ALiBi Slope Schedule}
\label{app:alibi}

ALiBi assigns each head a slope $m_h = 2^{-8(h+1)/H}$ controlling the distance penalty. For BLOOM-1b7 ($H = 16$):

\begin{table}[H]
\centering
\caption{ALiBi slope values and distance penalties for BLOOM-1b7 ($H=16$).}
\label{tab:alibi}
\begin{tabular}{crrr}
\toprule
Head & Slope & Distance Penalty at pos 10 & At pos 100 \\
\midrule
H0  & 0.7071 & $-7.07$ & $-70.71$ \\
H4  & 0.1768 & $-1.77$ & $-17.68$ \\
H8  & 0.0442 & $-0.44$ & $-4.42$ \\
H12 & 0.0110 & $-0.11$ & $-1.10$ \\
H15 & 0.0039 & $-0.04$ & $-0.39$ \\
\bottomrule
\end{tabular}
\end{table}

The penalty is added to attention logits before softmax. At H15, the penalty for attending to position 100 is $-0.39$ nats---small in absolute terms, but sufficient to make position 0 (penalty $= 0$) consistently preferred during pretraining's many gradient steps. Slope values verified against the BLOOM implementation in HuggingFace Transformers (\code{build\_alibi\_tensor}).

\section{Reproducibility}
\label{app:repro}

\textbf{Hardware:} NVIDIA RTX 5070 Ti, 16GB VRAM. All experiments run on a single GPU.

\textbf{Software:} PyTorch 2.9.1, Transformers 4.57.1, CUDA 12.8.

\textbf{Random seeds:} C4 baseline uses seed=42 (set via \code{random.seed()}, \code{torch.manual\_seed()}, \code{torch.cuda.manual\_seed\_all()}). Pass 1 and Pass 2 used PyTorch defaults without explicit seed setting. We report results from the canonical runs; the 11 development iterations of Pass 1 showed consistent head recovery across all runs despite different random states.

\textbf{Data:} Stock BLOOM-1b7 weights from \code{bigscience/bloom-1b7} on HuggingFace. C4 validation split from \code{allenai/c4}. Curated corpus available at \url{https://github.com/Palmerschallon/bloom-head-surgery}. Surgical checkpoints available at \url{https://huggingface.co/TheNexus42/bloom-1b7-head-surgery}.

\textbf{Code:} Diagnostic tool, surgical training scripts, and evaluation code available at \url{https://github.com/Palmerschallon/bloom-head-surgery}.

\section{Supplementary Figures}
\label{app:figures}

\begin{figure}[H]
  \centering
  \includegraphics[width=0.85\textwidth]{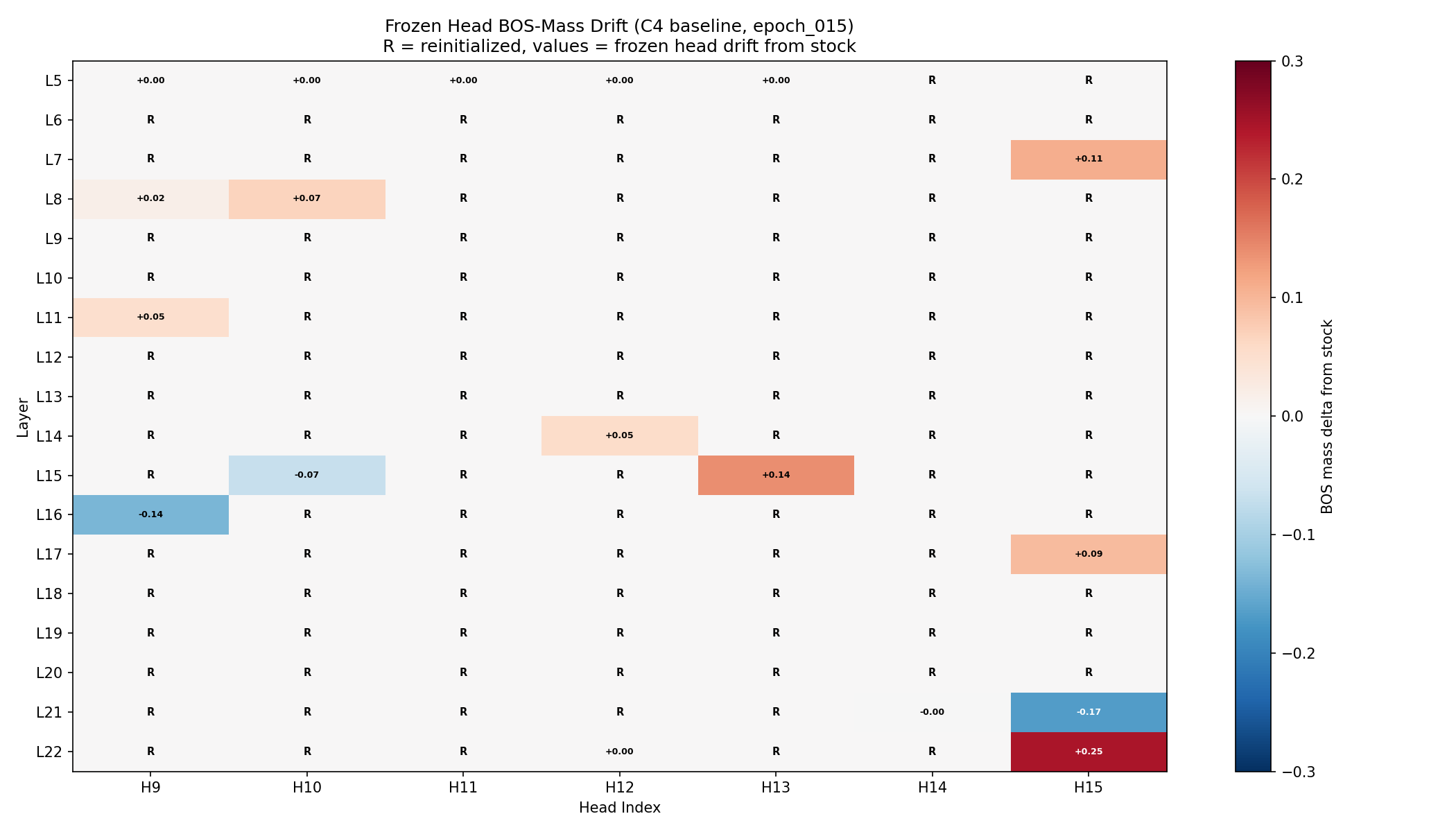}
  \caption{Frozen in-band head drift heatmap across training epochs (C4 baseline). Each cell shows the absolute BOS mass change $|\delta|$ for a frozen head (H9--H15 band, healthy at surgery time) relative to stock. Drift accumulates progressively in the H15 column while H12 and H10 remain largely stable, consistent with the column-specific propagation pattern described in Section~4.5.}
  \label{fig:frozen_drift}
\end{figure}

\begin{figure}[H]
  \centering
  \includegraphics[width=0.85\textwidth]{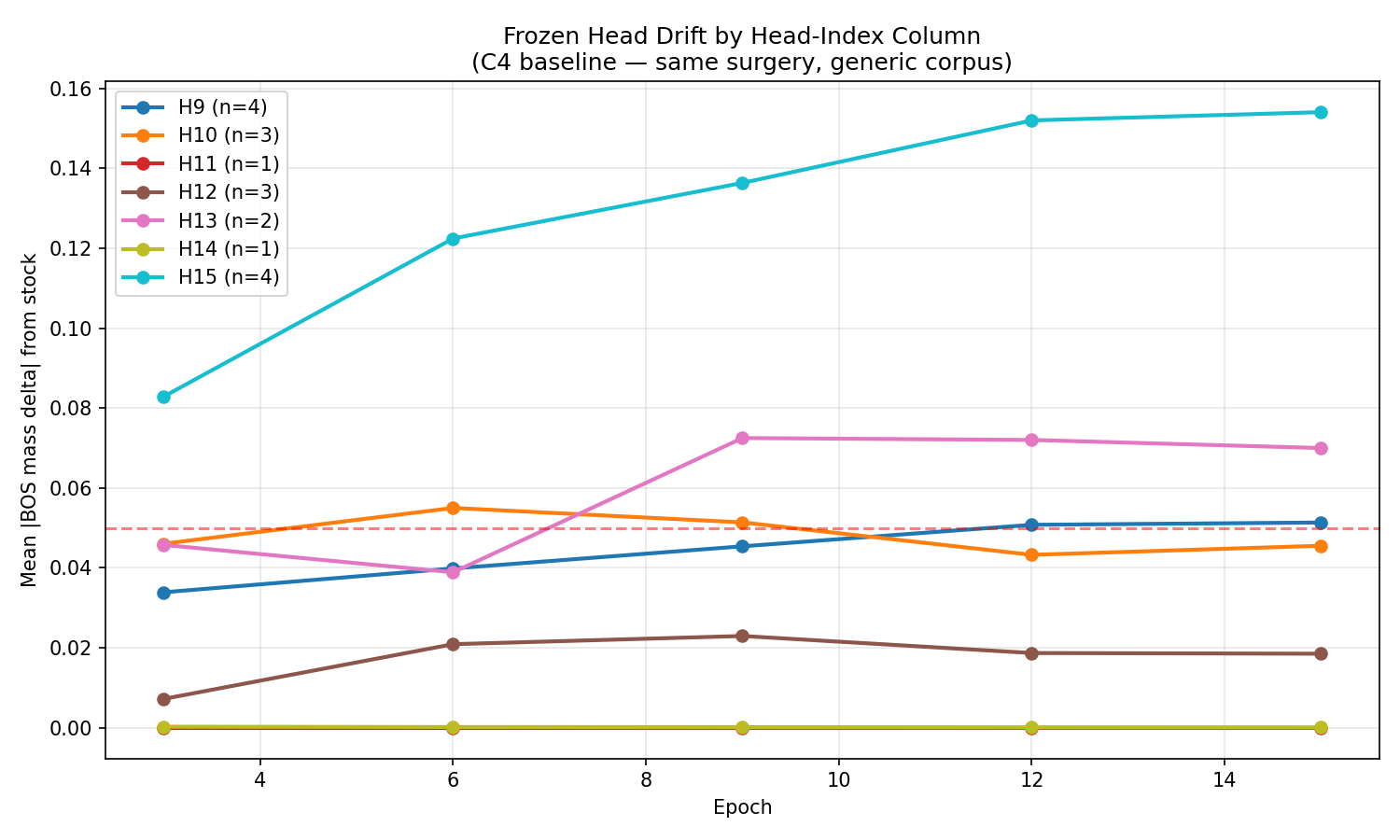}
  \caption{Column-wise drift trajectories across training epochs (C4 baseline). Each line tracks the mean $|\delta|$ for frozen heads within a given head-index column. H15 shows a spreading pattern (mean delta nearly doubles from E3 to E15), while H9, H10, and H12 remain stable. The differential vulnerability tracks the ALiBi slope schedule: steeper-slope columns are more susceptible to residual stream perturbation.}
  \label{fig:column_drift}
\end{figure}

\end{document}